# SLAM using ICP and graph optimization considering physical properties of environment

Ryuki Suzuki, Ryosuke Kataoka,
Yonghoon Ji, and Kazunori Umeda
*Department of Precision Mechanics,
Chuo University*
Tokyo, Japan
Email: r.suzuki@sensor.mech.
chuo-u.ac.jp

Hiromitsu Fujii
*Department of Advanced Robotics,
Chiba Institute of Technology*
Chiba, Japan

Hitoshi Kono
*Faculty of Engineering
Tokyo Polytechnic University*
Kanagawa, Japan

*Abstract*—This paper describes a novel SLAM (simultaneous localization and mapping) scheme based on scan matching in an environment including various physical properties. In scan matching, localization is performed mainly focusing on the shape information of the environment. However, the localization cannot be performed correctly and matching may fail when a similar shape existing in different places if only the shape information is taken into account. Therefore, we propose a new method to improve the accuracy of scan matching by considering abundant physical features existing in the environment. In this method, it is possible to utilize not only the shape information but also the physical information of the environment as features by measuring such as heat and water sources. Experiment result in the real environment shows that a highly accurate map of the environment can be generated by utilizing this physical information.

## I. Introduction

In Japan, which frequently experiences earthquakes, the development of robot systems in places where it is difficult for a human to enter is actively performed. However, many robots for the decommissioning activities were not able to return in the Fukushima Daiichi nuclear power plant station, which was damaged by the Great East Japan Earthquake in 2011. This is thought to be due to an inadequate understanding of the situation inside the nuclear power plant and lack of collected information. Therefore, it is important to develop a system to grasp the internal situation for the decommissioning activities. Inside the nuclear power plant station, it is expected that physical properties such as water leaked from the accident, heat sources from the explosion, and radiation will be distributed.

One of the general methods for understanding the situation is to generate an environmental map. In order to generate an environmental map using feature information extracted from sensor data, studies on SLAM (simultaneous localization and mapping) have been actively conducted. Mur-Artal et al. extracted ORB features from camera images and generated an environmental map [1]. Konolige et al. proposed a map generation scheme using the shape information of the measurement data from the LiDAR sensor (laser imaging detection and ranging) [2]. Hara et al. exploited scan matching method using laser reflection intensity from the LiDAR sensor to generate the map information [3]. However, these studies only deal with the shape information and one additional type of feature information such as color or intensity data. Thus, it is difficult to fully use the abundant physical properties in the nuclear power plant station described above. In this study, we propose a novel map generation scheme by scan matching using environmental features such as heat sources and a puddle of water in the nuclear power plant stations. By assigning such environmental features to the map information, it is possible to visualize hazardous areas from the heat sources and a puddle of water, which can be useful for investigating reactor buildings.

## II. Related Research

In SLAM, a method based on laser scan matching is widely used. A typical method to carry out scan matching is ICP (iterative closest point) [4]. ICP is highly extensible, and there are many extended versions [5,6,7]. In this study as well, we aim to improve the performance of SLAM by exploiting the environmental features in the nearest neighbor search and optimization process of ICP. There are also many approaches that extract features from camera images for SLAM [8,9]. However, it is difficult to obtain environmental features because the visibility of the camera is poor in the case of the damaged reactor building. There are several studies that utilize the semantic information of point clouds in SLAM. Civera et al. utilized recognized objects from a set of measurement points by template matching with objects in a database to improve SLAM performance [10]. Sunderhauf et al. incorporated the distinction between recognized object types and orientations into matching [11]. Mozos et al. reconstructed a dynamic environment based on SLAM taking into consideration of human activity area [12]. However, these studies focus only on the shape and color information of point clouds and do not consider the physical properties of the environment.

## III. Overview

In this study, we utilize physical property information of an environment such as water and heat sources as features. Depending on the type of sensor that integrates with the LiDAR sensor used for environmental measurement, the point cloud can include various information in addition to the 3D (three-dimensional) shape information of the surrounding environment and the reflection intensity of the laser. For example, integrating optical camera images, it is possible to obtain colored point cloud data. Near-infrared camera, thermal camera, and gamma camera also can be integrated in order to acquire physical property information of an environment such as water, heat, and radiation sources, respectively. In this study, a heat source class, a puddle of water class, and a class that does not belong to them are assigned to the 3D point cloud obtained from the LiDAR sensor. We propose a method to improve the alignment result in scan matching by considering class information for each point.

Figure. 1 shows the overall process of the proposed method. First, the robot explores the environment by remote control and obtains the point cloud data that include 3D shape information with physical properties for each frame. Next, an initial position for the ICP algorithm is calculated between the point clouds at *t*-1 frame and *t* frame by using odometry. Then scan matching is carried out based on the proposed ICP algorithm taking the physical properties into consideration to calculate the rotation matrix $R$ and translation vector $T$ between each frame. $R$ and $T$ are the parameters of the rigid transformation for aligning the point clouds in each frame. Considering physical properties such as water and heat sources, points with the same physical property can be more easily matched and mismatching can be reduced, compared to the conventional ICP algorithm. Next, by applying $R$ and $T$ of each frame to the robot pose, the localization is performed sequentially. When calculating the ICP algorithm with the physical properties, the covariance matrix for the robot pose is also calculated simultaneously. This covariance matrix represents how overlapping point clouds are; therefore, it contains the uncertainty of the robot pose in each frame.

Although the ICP algorithm minimizes the matching error between point clouds, the error does not completely disappear and the generated map is misaligned because pose errors accumulate over time. Therefore, we use the environmental features to detect that the robot has returned to the same place, and optimize the robot trajectory based on pose graph optimization. Here, the information matrix, which is the inverse of the covariance matrix is used to correct the robot trajectory. Since the calculated covariance includes the physical properties of the environment, these are also taken into account when performing the pose graph optimization. Therefore, more accurate pose graph optimization can be carried out.

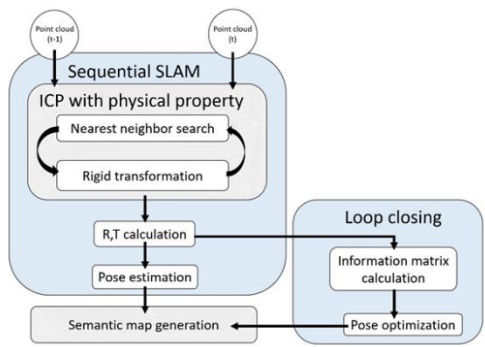

Fig. 1 Overview of proposed SLAM framework.

### IV. ICP WITH PHYSICAL PROPERTY

*A. ICP Algorithm*

In this study, we propose a method based on the ICP algorithm mainly used in scan matching-based SLAM. Here, the conventional ICP algorithm is briefly described. First, the initial pose of each point cloud for alignment is set. In the case of SLAM using a mobile robot, the estimated pose of the mobile robot based on odometry is generally set as the initial pose. Next, each point cloud is corresponded by the nearest neighbor search and aligned. After that, we solve the minimization problem that evaluates the alignment by the sum of the distances between point clouds. Here, each of successive point cloud is defined as the source and target point clouds, respectively. The rotation matrix $R$ and the translation vector $T$, are obtained by the minimization problem. The evaluation function for the minimization is as follows.

$$E = \sum_{i=1}^{N} \left| p_{k_i} - (q_i R + T) \right|^2 \qquad (1)$$

$E$: Sum of squares of distance (evaluation value)
$p$: Point of source point cloud
$q$: Point of target point cloud
$N$: Number of points in the source point cloud
$k_i$: Index number corresponding to *i*-th point in the source point cloud
$R$: Rotation matrix of the transformation matrix
$T$: Translation component vector of the transformation matrix

*B. ICP with physical property*

In this subsection, we describe the proposed ICP algorithm with physical properties. The calculation process of the matrices for $R$ and $T$ is the same as the conventional ICP; however, the physical properties are taken into consideration when searching the nearest neighbors. In general, the number of points including physical property is smaller than those without; thus, it is highly likely that these are measured from the same area in the environment. In other words, point clouds with the physical property are very useful as features for registration. Here, in order to search the same physical property by expending a scope of the nearest neighbor search in case of points with physical property, these can be easily aligned, as shown in Fig. 2. Modified evaluation function taking the physical properties of the point cloud into consideration can be written as follows:

$$E = \sum_{i=1}^{N} \left| p_{k_i} - (q_i R + T) \right|^2 + \lambda(p_{k_i}, q_i)$$

$$\lambda(p, q) = \begin{cases} \alpha & \text{if } c(p) \neq c(q) \\ 0 & \text{if } c(p) = c(q) \end{cases} \qquad (2)$$

where $\alpha$ indicates penalty given to the Euclidean distance to avoid physical property mismatch. The function $c(p)$ means the physical property on the point of the source point cloud.

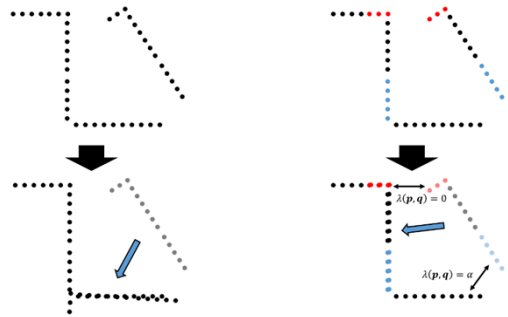

(a) Conventional ICP      (b) Proposed ICP

Fig. 2 Conceptual image of ICP algorithm.

## V. POSE GRAPH OPTIMIZATION

In the sequential ICP-SLAM with the physical property, it is impossible to completely eliminate the trajectory error of the robot because the measured point cloud contains errors. Furthermore, since this error accumulates for each frame, when the total number of frames increases, the accuracy of the localization is greatly affected. Therefore, we utilize physical properties as landmark information. In other words, the physical properties measured at different frames are compared and used for matching. In order to carry out the matching process mentioned above, we use pose graph optimization. As shown in Fig. 3, The pose graph represents the robot trajectory in a graph structure as nodes and the relative pose between nodes as edges. The residual $J$ of the pose graph up to the latest frame is represented as follows:

$$J = \sum_{t}\bigl(f(x_{t-1}, x_t) - d_{t-1,t}\bigr)^T \Sigma_t^{-1}\bigl(f(x_{t-1}, x_t) - d_{t-1,t}\bigr) \\ + \sum_{(s,t)\in C}\bigl(f(x_u, x_t) - d_{u,t}\bigr)^T \Sigma_{u,t}^{-1}\bigl(f(x_u, x_t) - d_{u,t}\bigr) \quad (3)$$

where $f(x_{t-1}, x_t)$ means the state transition model from $x_{t-1}$ to $x_t$. Here, $x$ denotes the robot pose and $d_{t-1,t}$ is the relative pose of the edge between nodes. $u$ is the frame observing the landmark at $t$. $\Sigma_t$ is the covariance matrix at frame $t$, which represents the uncertainty the robot pose at $t$, and $\Sigma_{u,t}$ represents the uncertainty of the relative pose between $t$ and $u$ frames. In addition, this covariance is calculated when ICP with physical property is executed for $p$ and $q$; thus, the covariance matrix contains physical property information. By calculating the combination of nodes that minimizes $J$, the robot trajectory can be optimized.

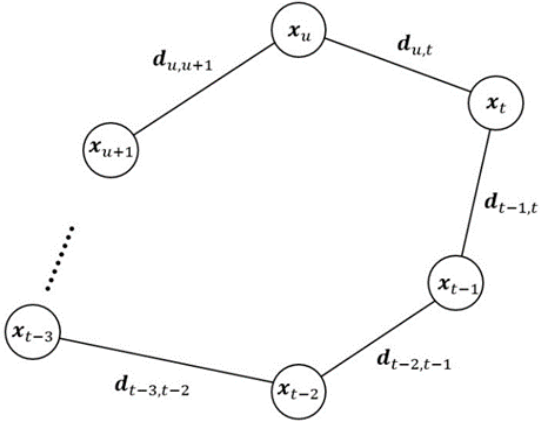

Fig.3 Pose graph generation.

## VI. EXPERIMENT

### A. Experiment condition

The map information of the environment generated by the LiDAR was used to evaluate the method. We measured the environment using an exploration robot equipped with the LiDAR by remote control. Figure. 4 shows the experimental equipment used in this study. Table 1 shows the specifications of the exploration robot. The exploration robot is equipped with Velodyne LiDAR VLP-16 and has an encoder that calculates the amount of movement from the amount of wheel rotation. ICP-SLAM was performed using the odometry obtained from the encoder as the initial position of scan matching. The experiment was performed in the Naraha Remote Technology Development Center shown in Fig. 5. In this facility, there are abundant buildings with characteristic shapes such as mock-up stairs and water tanks, thus, this facility is suitable for conducting ICP-SLAM evaluation experiments. In this experiment, an environment with physical properties was represented by assigning virtual physical property classes to point clouds obtained from LiDAR. Figure 6 shows a map generated using true robot trajectories. Red and blue parts of the map are the virtually simulated heat sources and the puddle of water, respectively. The purple line represents the true robot trajectory.

Table 1 Specification of exploration robot.

| Uphill slope angle [deg] | 45 |
|---|---|
| Payload [kg] | 5 |
| Maximum speed [mm/s] | 100 |
| Length [mm] | 1000 |
| Width [mm] | 400 |
| Height [mm] | 200 |

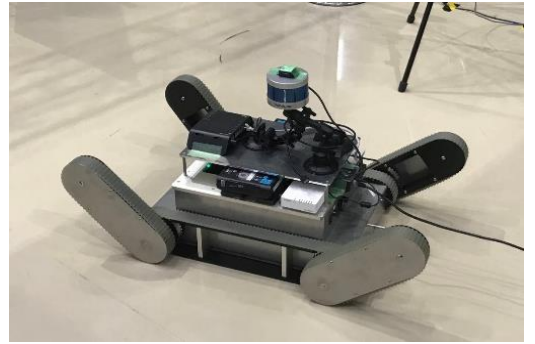

Fig. 4 Exploration robot.

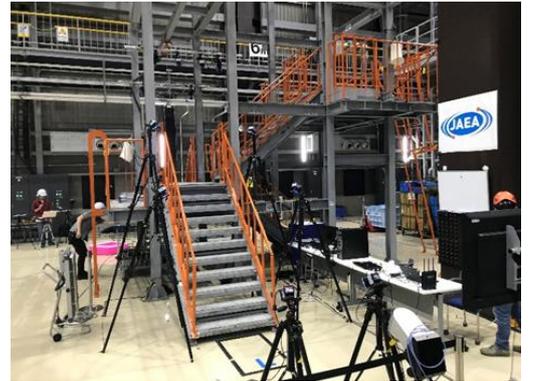

Fig. 5 Experimental environment.

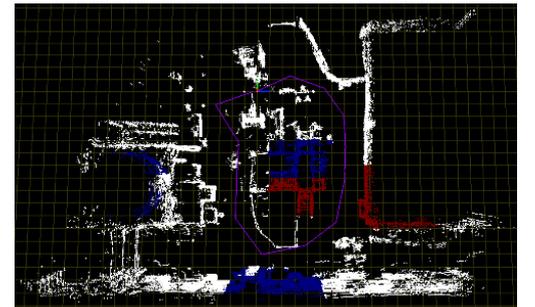

Fig. 6 Map built by true trajectory.

## B. Experiment result

As shown in Fig. 7, all the methods using ICP with physical properties reduced errors in the robot trajectory. In addition, the method that uses ICP with the physical property and pose graph optimization had the smallest error. On the other hand, in the method using conventional ICP and pose graph optimization, the error of the final frame was reduced, but the error did not decrease much on the whole. Pose graph optimization is intended to correct small errors that accumulate in sequential ICP-SLAM. However, in this experiment, the pose graph optimization was not performed well because the accumulated error by the conventional sequential ICP-SLAM was too large. In addition, the reason why the accumulated error of the conventional ICP-SLAM was large is that the error of the initial robot pose due to odometry was large. Generally, the odometry error of a crawler type robot is large. In this experiment, the error of the initial robot pose due to the odometry became large because the crawler type robot was used and the movement of the robot in each frame was large. From Figs. 8-11, the accuracy of the map was improved in each case using the proposed ICP compared to the conventional ICP. Table 2 shows that the error of the point cloud of the map generated by the method using the proposed ICP and pose graph optimization was the smallest, and the accuracy of the map was improved.

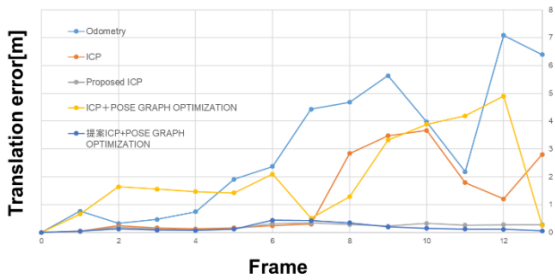

Fig. 7 Comparison of translation errors.

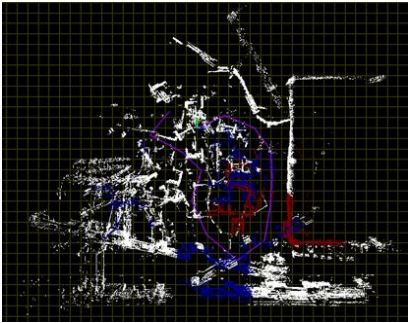

Fig. 8 Map by ICP.

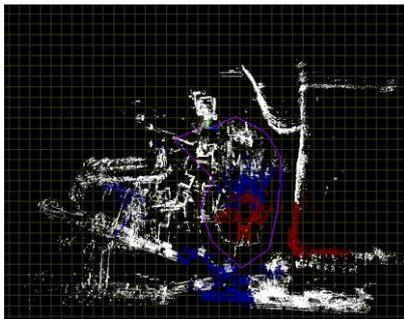

Fig. 9 Map by ICP with pose graph optimization.

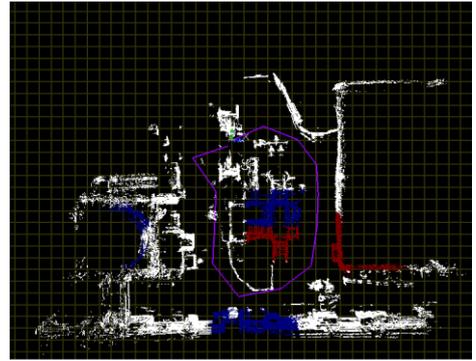

Fig. 10 Map by proposed ICP.

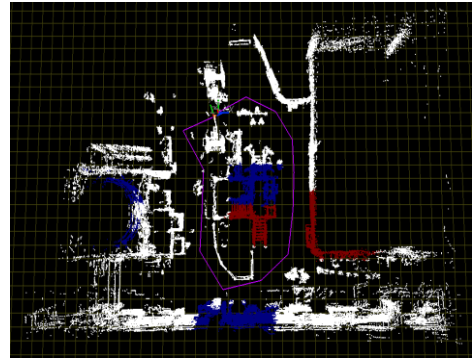

Fig. 11 Map by proposed ICP with pose graph optimization.

Table 2 Distance to the map build by true trajectory.

| Method | Error |
|---|---|
| Odmetry [m] | 0.918 |
| ICP [m] | 0.320 |
| ICP with pose graph optimization [m] | 0.322 |
| Proposed ICP [m] | 0.023 |
| Proposed ICP with pose graph optimization [m] | 0.011 |

## VII. CONCLUSIONS

In this study, we proposed a method to generate an environmental map by applying physical properties to ICP. In the proposed method, the SLAM system that considers physical properties was constructed by widening scope of the nearest neighbor search and providing a penalty when matching different physical property points in the ICP algorithm. In the experiment, the alignment was more accurately performed by the ICP with physical property than the conventional method, and a highly accurate map was generated.

## ACKNOWLEDGMENT

This work was founded by The Japan Atomic Energy Agency (JAEA), The Center of World Intelligence Project for Nuclear Science and Technology and Human Resource Development (Grant no. 30I107).